\documentclass[letterpaper, 10 pt, conference]{ieeeconf}  
\usepackage{amsmath,amsfonts}
\usepackage{mathrsfs}
\usepackage{textcomp}
\usepackage{stfloats}
\usepackage{url}
\usepackage{cite}
\usepackage{amssymb}
\usepackage[dvipsnames]{xcolor}
\usepackage{balance}
\usepackage{graphicx}
\usepackage{color}
\usepackage{svg}
\usepackage{xfakebold}
\newcommand{\fbseries}{\unskip\setBold\aftergroup\unsetBold\aftergroup\ignorespaces}
\newcommand{\setBoldness}[1]{\def\fake@bold{#1}}
\usepackage{listings}

\definecolor{codegreen}{rgb}{0,0.6,0}
\definecolor{codegray}{rgb}{0.8,0.8,0.8}
\definecolor{commentgrey}{rgb}{0.5,0.5,0.5}
\definecolor{backcolour}{rgb}{0.98,0.98,0.98}
\definecolor{functionblue}{rgb}{0.1,0.1,0.7}
\definecolor{box}{cmyk}{0.0,0.0,0.2,0.0}

\lstdefinestyle{myPython}{
    backgroundcolor=\color{backcolour},   
    commentstyle=\color{commentgrey},
    keywordstyle=\color{codegreen},
    numberstyle=\tiny\color{codegray},
    basicstyle=\ttfamily\footnotesize\fbseries,
    morekeywords={as},
    breakatwhitespace=false,         
    breaklines=true,                 
    captionpos=b,
    emph={policy, policy_v0, policy_v1, policy_v2, evaluate},
    emphstyle={\color{functionblue}},
    keepspaces=true,                 
    numbers=none,                    
    numbersep=5pt,                  
    showspaces=false,                
    showstringspaces=false,
    showtabs=false,                  
    tabsize=2
}

\newcommand{\blue}[1]{\textcolor{black}{#1}}

\IEEEoverridecommandlockouts                              

\makeatletter
\let\NAT@parse\undefined
\makeatother
\usepackage{hyperref}
\hypersetup{
  colorlinks=true,
  linkcolor=red,
  urlcolor=blue,
  citecolor=green,
}
\title{\LARGE \bf
Synthesizing Interpretable Control Policies through Large Language Model Guided Search
}
\author{Carlo Bosio and Mark W. Mueller
\thanks{The authors are with the High Performance Robotics Laboratory, University of California, Berkeley. Contact: {\tt\small \{c.bosio, mwm\}@berkeley.edu}}
}
\begin{document}
\maketitle
\thispagestyle{empty}
\pagestyle{empty}
%
\begin{abstract}
The combination of Large Language Models (LLMs), systematic evaluation, and evolutionary algorithms has enabled breakthroughs in combinatorial optimization and scientific discovery. We propose to extend this powerful combination to the control of dynamical systems, generating interpretable control policies capable of complex behaviors. With our novel method, we represent control policies as programs in standard languages like Python. We evaluate candidate controllers in simulation and evolve them using a pre-trained LLM. Unlike conventional learning-based control techniques, which rely on black-box neural networks to encode control policies, our approach enhances transparency and interpretability. We still take advantage of the power of large AI models, but only at the policy design phase, ensuring that all system components remain interpretable and easily verifiable at runtime. Additionally, the use of standard programming languages makes it straightforward for humans to finetune or adapt the controllers based on their expertise and intuition. We illustrate our method through its application to the synthesis of an interpretable control policy for the \textit{pendulum swing-up} and the \textit{ball in cup} tasks. We make the code available at \url{https://github.com/muellerlab/synthesizing_interpretable_control_policies.git}.
\end{abstract}
%
%
\section{INTRODUCTION}
\begin{figure}[!tb]
\vspace{0.2cm}
    \centering
    \includegraphics[width=0.8\linewidth]{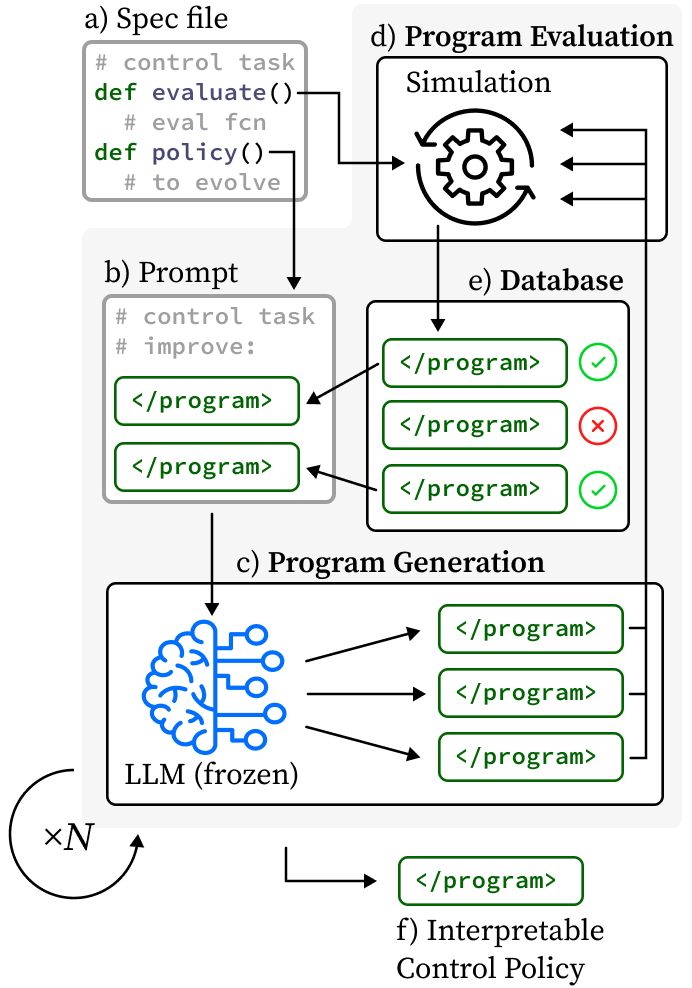}
    \caption{Schematic of the algorithmic infrastructure for the synthesis of interpretable control policies. The input to the algorithm is a specification file a) containing a task description, the implementation of an evaluation function to score programs, and some starter code for the control policy to evolve. A prompt b) is constructed pasting the current best programs (the starter code at the beginning). The prompt is fed to a \textit{Program Generation} block c) containing a pre-trained LLM, which produces more programs. The control policies contained in the LLM outputs are fed to the \textit{Program Evaluation} block d), which scores them based on their performance in simulation. The programs leading to poor performance are discarded, while the higher scoring ones are stored in a \textit{Database} e), from which they are sampled to be included in following prompts and improved.}
    \label{fig:block-diag}
    \vspace{-0.4cm}
\end{figure}
Control systems and artificial intelligence (AI) are two fields with immense practical impact, yet their integration often faces significant challenges. While control theory offers reliable methods to stabilize and steer complex systems, recent advances in machine learning (ML) have dramatically improved our ability to leverage large-scale data. However, the use of black box AI models, particularly neural networks, is not always suitable for critical control applications where transparency and verifiability are essential. Our work introduces a solution to this problem. Inspired by recent breakthroughs in Combinatorial Optimization \cite{romera2024mathematical}, we propose representing control policies as programs written in standard languages like Python, and evolving them using a pre-trained LLM and a simulation framework for evaluation. 
Our approach still leverages the power of large AI models, but shifts the abstraction layer, moving the black box component from the runtime execution to the policy design phase (an outline of our method is shown in Fig. \ref{fig:block-diag}). The output of our control synthesis framework is a fully interpretable program representing a control policy for the system and task of interest. The key advantage of our approach is the use of code as the policy representation. Programming languages, being our primary tools for instructing machines, are inherently interpretable. A control engineer or system operator can read, understand, and even modify the policy directly, without needing to decipher complex neural network architectures or weight matrices. 

In the following sections we frame in more detail this research by highlighting relevant previous work. \blue{We first present an overview of related works in Section \ref{sec:related-work}.} In Section \ref{sec:method} we describe our methodology and infrastructure. Then, in Section \ref{sec:casestudies} we show an application of the proposed method and in Section \ref{sec:conclusion} we provide a brief discussion and conclude by highlighting possible future directions.


%
%
\section{\blue{Related Work}\label{sec:related-work}}
\blue{In the following we present some of the previous work which motivated the development of this idea.}
\subsection{Large Language Models}
Recent developments in Large Language Models (such as \cite{achiam2023gpt}) have opened novel research areas focused on their applications. One popular application is LLMs for code. In fact, a lot of effort has been put into developing, training, and finetuning large models for code generation \cite{fried2022incoder, li2023starcoder, lozhkov2024starcoder}. With the rise of these models for code, also evaluation and benchmarking have been popular research directions \cite{austin2021program, chen2021evaluating}. On top of LLMs for code generation, extensive system-level research has been proposed with the aim of integrating LLMs with additional components to achieve more complex tasks than with single prompt engineering. A popular example is the combination of LLMs with evaluators, i.e. programmatic ways of scoring their outputs. This paradigm has been applied in various contexts, such as automated reasoning \cite{zelikman2022star}, code debugging \cite{haluptzok2022language}, and algorithm design \cite{ lehman2023evolution, liu2024example}. The integration of this generation-evaluation technique in an evolutionary procedure alleviates LLM hallucination, and in some cases leads to self-improving loops which output high performance programs and new knowledge \cite{romera2024mathematical}. It is important to highlight that these results are not achieved thanks to domain-specific knowledge contained in the LLM training dataset. Instead, they are achieved through the combination of the capability of LLMs to generate functional code, and evolutionary optimization techniques.
Our work shows how to effectively apply these LLM-based frameworks to control systems design.
\subsection{Learning-Based Control}
Control systems have benefited in many ways from learning-based techniques \cite{hou2013model}. 
A popular and well-studied field is adaptive control, which revolves around the identification of discrepancies between a system's dynamics model and its real-world behavior, and update of model parameters to compensate this mismatch \cite{hovakimyan2010L1, sastry2011adaptive}.
Learning-based methods have found successful application in highly repetitive scenarios, where error patterns are iteratively approximated and incorporated in the controller over multiple executions of the same task. These techniques fall under the iterative learning control category \cite{bristow2006survey}.
Another family of methods, typically referred to as imitation learning, consists of approximating a control policy for a given task through a set of expert demonstrations (i.e. trajectories accomplishing the task of interest) \cite{hussein2017imitation}.
Reinforcement learning (RL) represents a significant shift in control paradigms, offering a way to approximate optimal control policies through interaction with the environment \cite{sutton2018reinforcement}. Recent advances in RL have led to important results in areas where traditional methods fall short, such as locomotion \cite{hwangbo2019learning, radosavovic2024real} and manipulation \cite{kalashnikov2018scalable}. 
However, one fundamental issue preventing all these techniques to be safely and reliably deployed in real world contexts is their lack of interpretability.




%
\subsection{Interpretability in Learning-Based Control}
In the context of Machine Learning systems, interpretability is defined as the \textit{ability to explain or to present in understandable terms to a human} \cite{doshi2017towards}.
This property is particularly crucial for control systems and automation, where guaranteeing safety of operation is in many cases essential. In such safety-critical contexts, being able to inspect and understand which building block of a system led to a failure is of fundamental importance, and is not possible with black box components such as neural networks. Interpretability has been investigated through several different approaches \cite{glanois2024survey}. One of them is the use of interpretable architectures for learning-based control. Some examples are decision trees \cite{paleja2023interpretable} and fuzzy controllers \cite{hein2018generating, hein2020interpretable}.
These techniques facilitate the tracing of the decision-making process, providing some level of insight into the system's behavior.
Another approach that has been proposed to make learned components more interpretable is encouraging sparsity in parameter or weight matrices \cite{chu2020discovering}, as sparsity is often used as a proxy for interpretability \cite{rudin2022interpretable}.
We argue that interpretability in learning-based control should encompass not only the ability to read and (partially) understand the system's logic, but also the capacity to modify it with a clear understanding of the consequences. By representing control policies in a standard programming language, we aim to take a step towards the intuitive understanding required for the practical, safe deployment of learning-based control systems.

\section{METHODOLOGY \label{sec:method}}
Our work addresses the problem of finding a high-performance control policy for a control task of interest. We first describe the problem fundamentals, following the formalism of \cite{bertsekas2019reinforcement}, and then the algorithmic aspects of our interpretable control synthesis approach.

We are concerned with the study of a discrete-time dynamical system with dynamics in the form
\begin{equation}
    x_{t+1} = f(x_t, u_t),
\end{equation}
where $t\in\mathbb{N}$ is the time index, $x_t\in\mathbb{R}^n$ is the state of the system at time step $t$, and $u_t\in U \subset \mathbb{R}^m$ is the control input. At each time step $t$, a stage reward
\begin{equation}
    r_t = g(x_t, u_t) 
\end{equation}
is incurred. The general objective is to find a control policy $u_t = h(x_t)$ to maximize the cumulative reward 
\begin{equation}
    R = \sum_{t=0}^T r_t, \label{eq:return}
\end{equation} 
where $T$ is a specified time horizon. It is well known that this problem is in general very complex, and in many cases only approximate solutions can be found. Our goal is to produce an approximate solution (i.e., a control policy  approximately maximizing the cumulative reward) that is interpretable. A typical approach would be to pick a functional representation for the control policy (e.g., linear feedback, neural network) and optimize its parameters to maximize the cumulative reward. To guarantee interpretability, in our setting we represent the function $h(\cdot)$ directly as a program $\texttt{policy}(\cdot)$ in Python. Therefore, our control policy is encoded as
\begin{equation}
    u_t = \texttt{policy}(x_t).
\end{equation}
The goal is then to find a high-performing control program $\texttt{policy}^*(\cdot)$ by approximating the solution to the following optimization problem:
\begin{equation}
\begin{aligned}
    \texttt{policy}^*(\cdot)= & \max_{\texttt{policy}(\cdot)} \,\,\, R &&&\\
     & \,\,\text{s.t.} \,\,\,\, x_{t+1} = f(x_t, u_t), \forall t\in\{0, ..., T\}&&\\
     & \,\,\,\,\,\,\,\,\,\,\,\,\, u_t = \texttt{policy}(x_t),&& 
\end{aligned}
\end{equation}
where $R$ is defined in eq. \ref{eq:return}.
In this formulation, the search does not happen in a parameter space encoding a mathematical structure (such as, for instance, a neural network), but directly in the space of programs.

This space is infinite-dimensional and too complex to search over (for instance, there is no length limit). To work around this complexity, we leverage a Large Language Model finetuned on code to generate candidate programs and explore the space. LLMs for code are trained on a large corpus of human-written code examples, hence they have a bias towards concise, human-readable code. 
The use of an LLM shifts the optimization to the space of tokens, i.e. the elements the LLM uses to decompose a string of text (tokens can be individual words, individual characters, or a mix of these). 
Searching in the token space is complex. To start, tokens are discrete, thus there is not a notion of gradient to guide the search, and the number of possible programs of a given length can be combinatorially large. Furthermore, randomly sampled tokens almost surely lead to programs that are syntactically incorrect and do not run. The use of an LLM alleviates these complications. 

The main benefit of representing a control policy as a program is the inherent interpretability of programming language, which is by design the way humans instruct machines. In the following, we provide more detail about how to algorithmically deal with this problem, and find good candidate  control policies in the form $\texttt{policy}^*(\cdot)$.

Our method is inspired by the work presented in \cite{romera2024mathematical}.
The overall algorithmic infrastructure is shown in Fig. \ref{fig:block-diag}. The input is a specification file, where a description of the task, some starter code for the function to evolve and the implementation of the evaluation function are provided. At each iteration, a \textit{Program Generation} block (containing the LLM) produces candidate control programs for the task of interest. The proposed control policies are then fed to a \textit{Program Evaluation} block, which simulates the underlying system in closed loop. The best performing programs are stored in a \textit{Database} and then fed back into the prompts for the subsequent iterations, where the LLM is instructed to improve upon the previously produced programs. In the following we explain in more detail the functioning of this framework.
%
\subsection{Specification}
The input to our control synthesis framework is a specification file (Fig. \ref{fig:block-diag}a). The file is composed by three main parts:
\begin{itemize}
    \item A description of the control task to solve, together with packages and libraries the LLM-generated code can use;
    \item An initial candidate control program in the form of starter code, which will be evolved across different iterations;
    \item The implementation of the evaluation function used to score candidate programs.
\end{itemize}
The different parts of the specification are parsed and used at different stages of the pipeline. The starter code is pasted in the initial prompt (Fig. \ref{fig:block-diag}b) and fed to the LLM in the preliminary stages, when better programs have not yet been produced. The evaluation function is used in the \textit{Program Evaluation} block (Fig. \ref{fig:block-diag}d) to quantify the performance of candidate programs. An example of a general specification structure for our control synthesis method is shown in Fig.~\ref{fig:spec}. 

%
\begin{figure}[tb]
    \centering
    \begin{lstlisting}[language=Python]
 """Finds a policy for the control task."""

 # import libraries needed
 import numpy as np

 # evaluation function for the control policy
 def evaluate() -> float:
   """Returns the reward."""
   environment = initialize_env()
   observation = environment.get_observation()
   total_reward = 0.0
   for _ in range(1000):
     action = policy(observation)
     reward, observation = environment.step(action)
     total_reward += reward
   return total_reward

 # function to evolve
 def policy(obs: np.ndarray) -> float:
   """Returns a control action."""
   action = np.random.uniform()
   return action
\end{lstlisting}
\vspace{-0.2cm}
    \caption{Example template for a control synthesis specification.}
    \label{fig:spec}
    \vspace{-0.4cm}
\end{figure}
\subsection{Prompt Construction}
The prompt (Fig. \ref{fig:block-diag}b) is a crucial component in the pipeline, as it triggers and steers the LLM generation. At each iteration, a prompt is constructed by concatenating two previously generated high-performing programs. During the initial phases of the algorithm execution, only the starter code (from the specification file) is available, and is directly pasted into the prompt. As the evolution progresses and higher-performing programs are generated, these  are used in the prompt instead of the starter code. The prompt also contains an instruction to the LLM to improve upon the policies provided. The sampled policies are inserted in the prompt in the form \texttt{policy\_v0}, \texttt{policy\_v1}, and the function header of the policy to generate (in the form \texttt{policy\_vx}) is appended at the end. An example of a general prompt structure is shown in Fig. \ref{fig:prompt}.
\begin{figure}[tb]
    \centering
    \begin{lstlisting}[language=Python]
 """Finds a policy for the control task.
    On every iteration, improve policy_v1 over the policy_vX methods from previous iterations.
 """
 import numpy as np

 def policy_v0(obs: np.ndarray) -> float:
   """Returns a control action."""
   action = np.random.uniform()
   return action
   
 def policy_v1(obs: np.ndarray) -> float:
   """Returns a control action."""
   action = 0.0
   return action

 def policy_v2(obs: np.ndarray) -> float:
   """ Improved version of 'policy_v1'."""

\end{lstlisting}
\vspace{-0.2cm}
    \caption{Example template for a prompt. The LLM generates a body for the provided function signature.}
    \label{fig:prompt}
    \vspace{-0.4cm}
\end{figure}
\subsection{Program Generation} 
A pre-trained LLM is the generation engine of the \textit{Program Generation} block (Fig. \ref{fig:block-diag}c), in which candidate control programs are sampled. The LLM is queried with a prompt containing two high-performing control programs generated in previous iterations, and is instructed to improve them. This step encourages the combination of ideas from previous programs, and is equivalent to a crossover in classical evolutionary algorithms. It is important to highlight that no additional training or finetuning is carried out, and the LLM is kept frozen across the execution of our method.

A number of hyperparameters affect the generation performances of an LLM. We found that, for our use case, the randomness of the LLM generation and the amount of repetition in the output sequence are most central. At each token generation step the LLM produces a vector containing a numerical score for every token in the vocabulary. For token $i$, there is a score $x_i$.
These scores are then normalized to obtain a distribution to sample from. The sampling probabilities are obtained as
\begin{equation}
    p_i = \frac{e^{x_i/T}}{\sum_i e^{x_i/T}},
\end{equation}
where $T$ is the LLM temperature. A low $T$ makes the token distribution narrow, a larger $T$ makes it more uniform. In our implementation we set $T=1$. The token generation is also steered by other hyperparameters. One of them is \texttt{top\_p} (with value within $[0,1]$), which sets the percentile of the token options to consider. If \texttt{top\_p} is 1, then the whole vocabulary is considered, otherwise a fraction is discarded and the probabilities $p_i$ are renormalized. We set \texttt{top\_p} to $0.95$. The \texttt{repeat\_last\_n} parameter sets the length (in number of tokens) of the sliding window used to check for repetition. We set \texttt{repeat\_last\_n} to 15 (i.e. the 15 previously generated tokens are not considered for sampling).
%
%
\subsection{Program Evaluation}
Candidate control programs are parsed from the LLM output and fed to the \textit{Program Evaluation} block (Fig. \ref{fig:block-diag}d). The evaluation consists of testing the programs and quantifying their performance using the evaluation function provided in the specification file. The candidate control policy is deployed in a simulation environment, in closed loop with the dynamics of the system of interest. The performance in simulation is quantified by a numerical score (in our case, the return of eq. \ref{eq:return}). 
The score is then associated to the candidate control program to make a program-score pair. Syntactically incorrect programs are discarded, while promising program-score pairs are stored in the \textit{programs database}, from which they are sampled to be added to the subsequent prompts, and evolved. The simulation is run inside a sandbox to prevent issues (syntactic or of other nature) contained the LLM-generated control policy to interrupt the outer optimization routine.
\subsection{Programs Database}
The generated high performing programs are stored in the \textit{programs database} (Fig. \ref{fig:block-diag}e). At each iteration, two programs are sampled to be integrated into the prompt and fed back to the LLM, which is instructed to provide higher performing variants. To discourage getting stuck in local optima, an island approach is implemented, where different instances of the program search are run in parallel, independently \cite{cantu1998survey}. Therefore, an equivalent number of program populations are stored and evolved. Periodically, the islands containing less promising populations are emptied and re-initialized with the best performing programs from other islands. When constructing a prompt, a two stage sampling procedure happens: first, an island is sampled, then programs contained within the selected island are sampled to be added to the prompt. More detail about the algorithm implementation can be found in \cite{romera2024mathematical}. In our case, we use $10$ independently evolved islands.
%
\section{SET UP AND CASE STUDIES\label{sec:casestudies}}
In the following we provide more detail about the practical aspects of the implementation of our method. We then introduce the control tasks that we found interesting for its application.
\subsection{Setup}
We run our control synthesis framework on a workstation equipped with an NVIDIA RTX 3090 GPU, which has enough memory to fully load the LLM for inference. As a simulation framework for program evaluation we use the open source simulator MuJoCo \cite{todorov2012mujoco}, through the well known DeepMind Control Suite \cite{tassa2018deepmind} library. As LLM for program generation we used an 8-bit quantized version \cite{jacob2018quantization} of \textit{StarCoder2-Instruct} \cite{li2023starcoder}, an open source $15$ billion parameter model finetuned on code generation from natural language instructions. We applied our method to the \textit{pendulum swing-up} and \textit{ball in cup} tasks included in the DeepMind Control Suite. A supplementary video showing examples from the case studies can be found at \url{https://youtu.be/7T7yRGya-q8}.
\subsection{Pendulum swing-up} 
The \textit{pendulum swing-up} task with input constraints is not easily solvable through a classical linear control. The pendulum has to accomplish a number of oscillations to accumulate enough energy, and then swing into the upright configuration. The maximum applicable torque is $1/6^{\text{th}}$ as required to lift it from motionless horizontal.
The dynamics equations, with the angle $\theta$ representing the deviation from the upright position, are
\begin{equation}
    \Ddot{\theta} - \frac{g}{\ell}\sin\theta + b \, \Dot{\theta} = u,
\end{equation}
where $g = 9.81\,\mathrm{m\cdot s^{-2}}$ is the gravitational acceleration, $\ell = 0.5\,\mathrm{m}$ is the length of the pendulum (massless) rod, $b = 0.4\,\mathrm{s^{-1}}$ is a normalized damping coefficient, and $u$ is a normalized torque input.
When simulated, the dynamics equations are discretized through a semi-implicit Euler method and a sampling time of $15\,\mathrm{ms}$ (details in \cite{todorov2012mujoco}). At each time step $t$ a reward $r_t$ is computed. The overall score of the candidate control policy is the summation of individual rewards, i.e. $R$ defined in eq. \ref{eq:return}.
For the \textit{pendulum swing-up} task, the stage reward is defined as
\begin{equation}
    r_t = \begin{cases}
  1 - \frac{|\theta_t|}{\pi} - 0.1\,|u_t| & \text{if} \,\, |\theta| > 0.5\\
  2 - \frac{|\theta_t|}{\pi} - 0.1\,|u_t| & \text{otherwise.}
\end{cases}
\end{equation}
This reward function led to the best results, and was shaped through trial and error. The goal is to reward configurations which are close to the upright position, and penalize the input. However, another common behavior arising is an indefinite uniform spinning after the first transient phase.
\begin{figure}[tb]
    \centering
    \begin{lstlisting}[language=Python]
 def policy(obs: np.ndarray) -> float:
   """Returns an action between -1 and 1.
   obs size is 3.
   """
   x = np.arctan2(-obs[1], obs[0])
   if abs(x) < 0.5:
      action = 5*x - 0.9*obs[2]
   else: 
      action = np.sign(obs[2])

   return action
\end{lstlisting}
\vspace{-0.2cm}
    \caption{Best performing control program generated for \textit{pendulum swing-up} with our technique. The proposed policy applies positive work when the pendulum is not within a certain angular threshold from the upright position. Otherwise, it switches to a linear controller. The control action is normalized within $[-1,1]$, therefore, in the initial phases, the control takes one of the two limit values, depending on the sign of the angular velocity.}
    \label{fig:pendulum-policy}
\end{figure}

To find the high scoring policies for the \textit{pendulum swing-up} task, 
our framework generated in the order of $10^4$ sample programs per individual run. One run in particular led to the control policy shown in Fig. \ref{fig:pendulum-policy}. As it is possible to observe, the policy is highly compact and interpretable. Rewritten in mathematical terms, it corresponds to
\begin{equation}
    u_t = \begin{cases}
  5 \, \theta_t - 0.9 \, \Dot{\theta}_t & \text{if} \,\, |\theta_t| < 0.5\\
  \text{sgn}(\Dot{\theta}_t) & \text{otherwise,}
\end{cases} \label{eq:pendulum-policy}
\end{equation}
where $\text{sgn}(\cdot)$ is the sign function. Unpacking the policy, it applies positive work whenever the pendulum is not within a certain angular threshold from the upright position. Otherwise, it is a linear feedback controller. A user could make changes to this policy, such as tuning the linear control gains, or making the torque input a smoother function of the angular velocity, and keep iterating with the LLM in the loop. We also note that a Lyapunov indirect (local) stability analysis could be trivially carried out on eq. \ref{eq:pendulum-policy}.

\begin{figure}[!tb]
    \centering
    \includegraphics[width=0.99\linewidth]{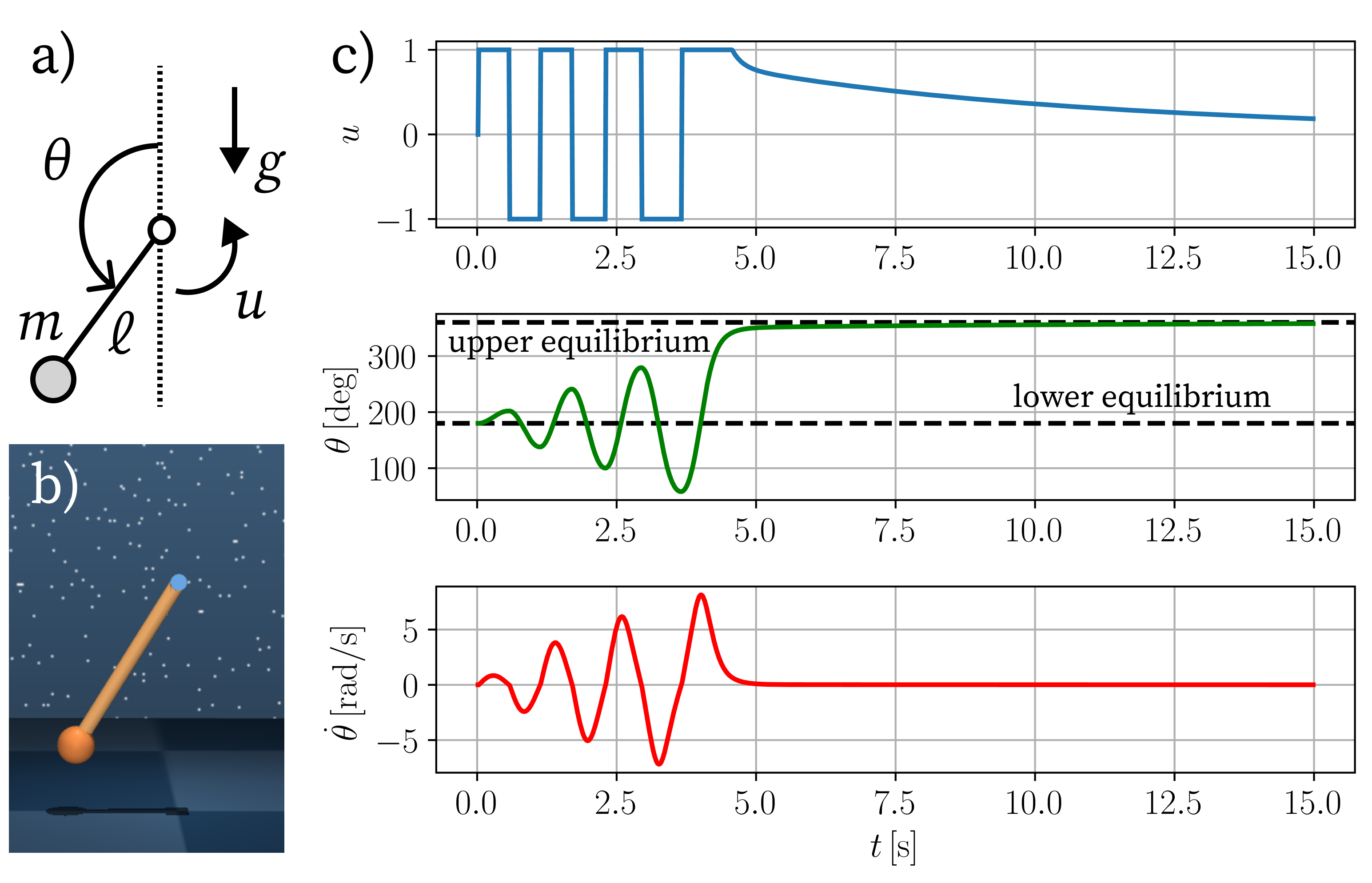}
    \caption{a) Schematic of the pendulum system and the angle convention. b) Screenshot of a visualization from the simulation environment. c) Example plots of the closed loop evolution for the \textit{swing-up} task. In the top graph, it is possible to observe a bang-bang style control in the first phase, followed by a linear feedback in the second phase.} 
    \vspace{-0.5cm}
    \label{fig:pendulum-plot}
\end{figure}

Plots of the closed-loop dynamics of the \textit{swing-up} task are shown in Fig. \ref{fig:pendulum-plot}. It is possible to observe that in the first stage the policy is of bang-bang type, and in a second stage it switches to a linear feedback.
\subsection{Ball in Cup}
\begin{figure}[!tb]
    \centering
    \includegraphics[width=0.99\linewidth]{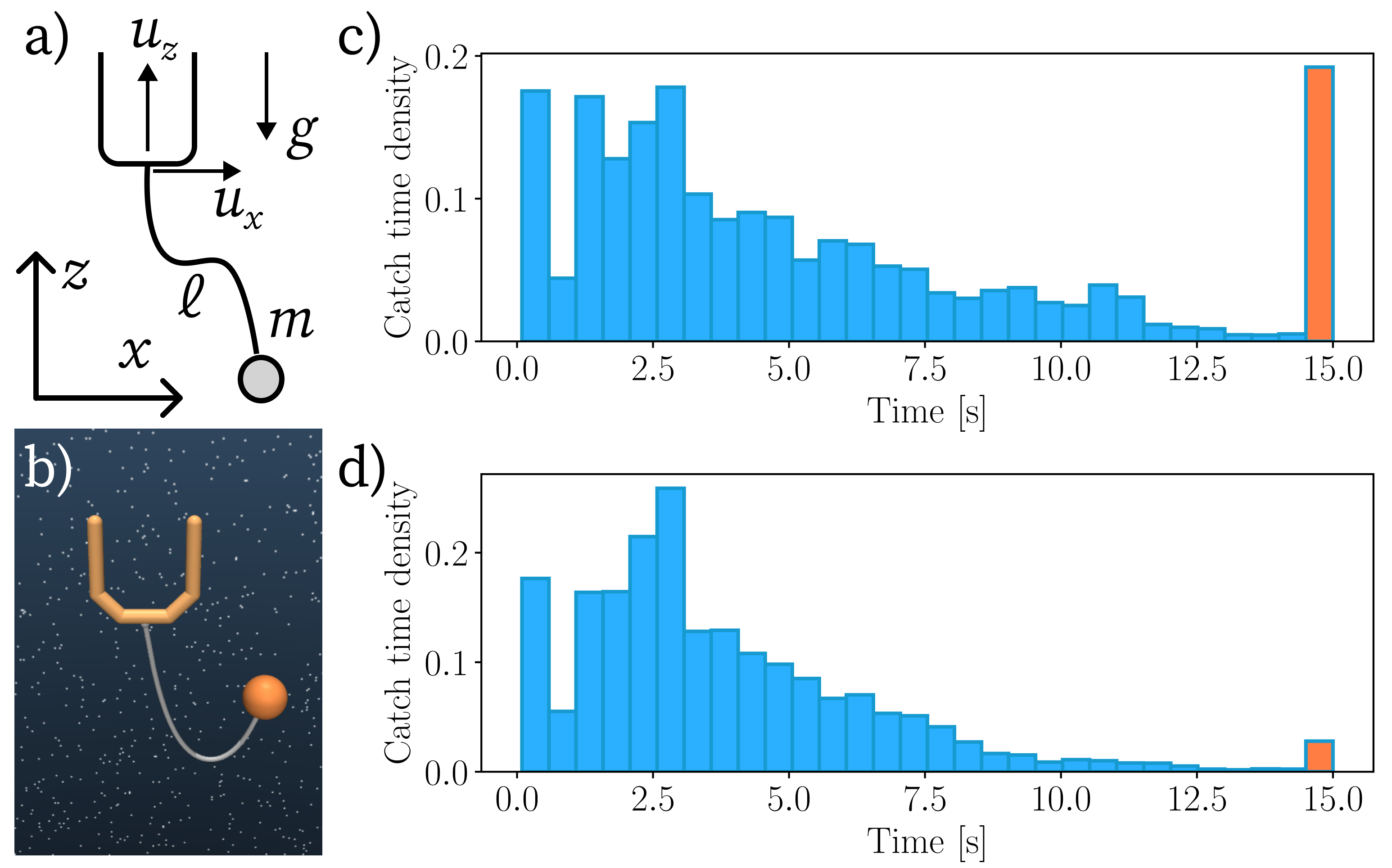}
    \caption{a) Schematics of the ball and cup system. $\ell = 0.3\,\mathrm{m}$,  $u_x$ and $u_z$ are the inputs to the independently actuated linear joints of the cup. b) Screenshot of a visualization from the simulation environment. c) Histogram showing the distribution of ball catching times across $10^4$ different episodes using the policy of Fig. \ref{fig:ballcup-policy}. At each episode, the cup is initialized at the origin of the task reference frame, and the ball position coordinates are sampled uniformly around the cup. Each episode is terminated when the ball is caught or after 15s. d) Histogram constructed with the same procedure using the user-improved policy. The two experiments use the same random seed. As it is possible to observe, the amount of episodes in which the ball is not caught within $15\,\mathrm{s}$ decreases significantly (orange bin).} 
    \label{fig:ballcup-hist}
    \vspace{-0.3cm}
\end{figure}

The \textit{ball in cup} system is composed by a two-dimensional double integrator (the cup) and a ball attached to it through a string (which is a unilateral distance constraint between the two entities). The system is planar. A schematic of the system is shown in Fig. \ref{fig:ballcup-hist}a. The task is to find a control policy for the cup (independently actuated along the horizontal and vertical direction) to catch the ball. In this scenario, the policy outputs a reference positions for the cup $(x_{ref}, z_{ref})$, which are then fed to a lower level linear controller. Therefore, this task is higher dimensional than the \textit{pendulum swing-up}, but does not involve any stabilization of the system.
For this task, the reward is defined as
\begin{equation}
    r_t = \begin{cases}
  1 - \frac{|\theta_t|}{\pi} - 0.1\,v_{ball} & \text{if ball outside cup}\\
  1 & \text{otherwise,}
\end{cases}
\end{equation}
where $x_{ball}$, $z_{ball}$, $x_{cup}$, $z_{cup}$  are the $x$ and $z$ coordinates of ball and cup (at time $t$, omitted for simplicity), $\theta_t = \mathrm{atan2}(x_{ball} - x_{cup}, z_{ball} - z_{cup})$ is an angle measuring how close the ball is to be vertically aligned with the cup, and $v_{ball} = \sqrt{\Dot{x}_{ball}^2 + \Dot{z}_{ball}^2}$ is the norm of the ball velocity (also at time $t$). Also in this case, the reward function was shaped through trial and error. 

The number of sampled programs is again in the order of $10^4$. The best found policy is shown in Fig. \ref{fig:ballcup-policy}, where the observation vector \texttt{obs} is $8$-dimensional, and organized as $(x_{cup}, z_{cup}, x_{ball}, z_{ball}, 
\Dot{x}_{cup}, \Dot{z}_{cup}, \Dot{x}_{ball}, \Dot{z}_{ball})$. 
At first, the policy is harder to parse compared to the \textit{pendulum swing-up} case. However, all the reasoning steps are clearly outlined and again, being a Python program, it is easy to modify and customize. To support this claim, we manually proceed to simplify the policy. First, the cup is constrained to the box $(x,z)\in S=[-0.25, 0.25]\times[-0.25, 0.25]$. Therefore, all the conditional statements checking for values outside this box (i.e. $0.8$) are never visited and can be removed. 
It is possible to also see from the first conditional statements, which assign $x_{ref}$ a value, that the policy commands a narrow motion of the cup along $x$ (roughly within $[0.2,0.25]$). In fact $x_{ref}$ only gets values of either $0.0$ if both $x_{cup}$ and $z_{cup}$ are above $0.2$, or $1$ otherwise (but the motion is constrained to $S$). Therefore, the policy mostly exploits a vertical stroke to swing the ball. The policy, cleaned of unused logic and with conditional statements grouped more meaningfully, is 
\begin{equation*}
    x_{ref} = \begin{cases}
  1 & \text{if} \,\, x_{cup} < 0.2\,\, \text{or}\,\, z_{cup} < 0.2 \\
  0 & \text{otherwise,}
  \end{cases}
\end{equation*}
\begin{equation}
  z_{ref} = \begin{cases}
      1 & \text{if}\,\,z_{ball} < 0.2 \,\, \text{or} \,\, \Dot{x}_{ball} > 0.5 \,\, \text{or} \,\, \Dot{z}_{ball} < -0.5\\
      -1 & \text{if}\,\, \Dot{x}_{cup} > 0.2 \,\, \text{or} \,\, \Dot{z}_{cup} < -0.2\\
      0 & \text{otherwise,}
  \end{cases}
\end{equation}
Additionally, we can leverage some intuition to even improve this policy. In fact, by visually inspecting the behavior of the system (a video can be found at \url{https://youtu.be/7T7yRGya-q8}), we noticed that a common failure is that the ball is not caught because it hits the sides of the cup. An intuitive modification would just be to insert a conditional statement that encourages the cup to lower slightly if the ball is swinging at a height (along $z$) which is greater than the cup's, to encourage catching.
This modification can be easily achieved through, for example, the following statement:
\begin{equation}
    \text{if} \,\, z_{ball} - z_{cup} > 0.1, \,\, \text{then} \,\, z_{ref} \leftarrow z_{ref} - 0.1,\label{eq:cond}
\end{equation}
which is equivalent to adding the following two lines of code as the last two lines of the policy in Fig. \ref{fig:ballcup-policy}:
\begin{lstlisting}[language=Python, escapechar=!]
    if obs[3] - obs[1] > 0.1:
       action[1] = action[1] - 0.1\end{lstlisting}
%
%
%
%
\begin{figure}[tb]
    \centering
    \begin{lstlisting}[language=Python, escapechar=!]
 def policy(obs: np.ndarray) -> np.ndarray: 
    """Returns two actions between -1 and 1.
    obs size is 8."""
    # x_cup, z_cup, x_ball, z_ball = obs[0:4]
    # vx_cup, vz_cup, vx_ball, vz_ball = obs[4:8]
    
    action = np.zeros((2,)) # x_ref, z_ref
    if obs[0] < 0.2:
      action[0] = 1
    elif obs[0] > 0.8:
      action[0] = -1
    if obs[1] > 0.8:
      action[0] = -1
    elif obs[1] < 0.2:
      action[0] = 1
    
    if obs[2] > 0.8:
      action[1] = 1
    elif obs[3] < -0.2:
      action[1] = 1
    elif obs[4] > 0.2:
      action[1] = -1
    elif obs[5] < -0.2:
      action[1] = -1
    
    if obs[6] > 0.5:
      action[1] = 1
    elif obs[7] < -0.5:
      action[1] = 1
      
    return action
\end{lstlisting}
%
%
%
\vspace{-0.2cm}
    \caption{Example policy found for \textit{ball in cup}. The commented parts were added after, to aid the reading of raw code. The output is in the form \texttt{action} $= [x_{ref}, z_{ref}]$ providing a reference position, the deviation from which is then consumed by an underlying low-level PD controller.}
    \label{fig:ballcup-policy}
    \vspace{-0.3cm}
\end{figure}
We evaluate the original policy of Fig. \ref{fig:ballcup-policy} and the user-improved policy (i.e. the raw policy with the additional condition of eq. \ref{eq:cond} at the end) by comparing their ball catching performances across $10^4$ different episodes. In particular, histograms showing the distribution of catching times for the raw policy and the user-improved one are shown in Figs. \ref{fig:ballcup-hist}c-d, respectively.
In each episode, the cup always starts at the origin of the task reference frame, and the ball position is uniformly sampled at random. The episodes are terminated either when the ball is caught, or when 15 seconds have passed. The two experiments were conducted a sufficient number of times to reach with acceptable approximation the convergence of histograms to the underlying distributions. The last bin of the histograms, in orange in Figs. \ref{fig:ballcup-hist}c-d, represents the number of episodes that terminated due to maximum time reached (and thus the ball was not caught). This shows that the catching rate improves significantly with the simple intuitive modification made. 
\section{DISCUSSION AND CONCLUSIONS \label{sec:conclusion}}
We presented a novel approach to synthesize interpretable control policies. The interpretability is inherently guaranteed by the representation of control policies as programs in standard programming language, and casting the problem as a program synthesis problem, whose solution is achieved with the code generation capability of a Large Language Model. The policy representation through code allows a user not only to read and understand the control system's logic, but also to have intuitive understanding of the effects of a manual modification to the program. This unlocks properties such as explainability, modularity, verifiability, and paves the way to joint iterative synthesis approaches where a user can actively steer the automatic search and collaborate with an LLM in-the-loop to get different policies based on the design requirements. \blue{User studies could help understand the effectiveness of our algorithm in satisfying these user-oriented metrics.}
\blue{A more rigorous formulation of how user feedback could be more formally integrated into the optimization loop is also an interesting future direction.}

The key reason for why this is possible is the shared language and formalism, i.e. Python in our case, between the user and the control system, as opposed to the use of black box models. Interpretability, however, comes with an increase in compute cost related to the absence of gradients to guide the optimization routine. 
In fact, as an example, our implementation running on a single GPU takes in the order of $10$ hours of wall-clock time to output programs able to successfully execute the proposed tasks. \blue{A multi-GPU implementation, as well as the use of lighter weight models, would reduce the wall-clock time.}

Another aspect worth discussing is the role of randomness in the LLM generation process. At every generation step, a token is sampled from a distribution over (almost) all possible tokens in the underlying vocabulary. This 
offers opportunities for further investigations on the robustness of our method, starting for example with a careful tuning of hyperparameters. 

Picking the right reward function, as in many reinforcement learning settings, is also crucial for success. Future works can focus on how to make the algorithm more computationally efficient, potentially incorporating gradient-based optimization in-the-loop (such as in \cite{ma2024llm}, for example). This would allow to take advantage of the LLM only as generator of a program skeleton, and unburden it from the task of producing the right numerical quantities by tuning them with continuous optimization techniques. \blue{This would also guarantee local control stability, which is not directly addressed in this work.} 

Compute availability is also an important aspect of the proposed framework. Implementing a distributed approach and running multiple LLM samplers in parallel allows to speed up the process and achieve high-performing programs in reduced time, as shown in \cite{romera2024mathematical}. \blue{More compute availability and improved sample efficiency would improve the scalability of the algorithm to higher-dimensional systems.}

It is also worth to mention
that the amount of domain-specific information provided in the specification can significantly impact the performances (in terms of runtime and sample efficiency) of the control synthesis procedure. 
LLMs, in fact, are in general sensitive to prompt variations. 
As a simple example, prompting an LLM with a generic plain instruction like ``generate a function'', compared to instead providing more details about the task of interest, can make a significant difference. Even if both approaches, down the line, could lead to high-scoring programs, it is typically beneficial to provide context information for sample efficiency. 

To conclude, we believe that code provides a compact, extremely expressive, and fully interpretable representation for a control policy. Thanks to these properties, we claim that our approach can reduce the gap between learning-based control systems and verifiable, reliable real world~applications. 
%
%
\section*{ACKNOWLEDGMENT}
This work was supported by the Hong Kong Center for Logistics Robotics (HKCLR), and the Powley fund of the Mechanical Engineering department of UC Berkeley. The authors would like to thank Orr Paradise for the insightful conversations.
%
%
%
\balance
\bibliography{IEEEabrv, refs}
\end{document}